\documentclass{article}

     \usepackage[preprint]{neurips_2022}

\usepackage[utf8]{inputenc} % allow utf-8 input
\usepackage[T1]{fontenc}    % use 8-bit T1 fonts
\usepackage{hyperref}       % hyperlinks
\usepackage{url}            % simple URL typesetting
\usepackage{booktabs}       % professional-quality tables
\usepackage{amsfonts}       % blackboard math symbols
\usepackage{nicefrac}       % compact symbols for 1/2, etc.
\usepackage{microtype}      % microtypography
\usepackage{xcolor}         % colors

\usepackage[colorinlistoftodos,size=small]{todonotes} % adapted by MZ

\usepackage[pdftex,outline]{contour}
\contourlength{0.3pt}% Thickness of copies

\usepackage{amssymb}
\usepackage{mathtools}
\usepackage{wrapfig}
\usepackage{caption}

\newtheorem{definition}{Definition}

\newtheorem{conjecture}{Conjecture}

\newcommand{\sol}{\mathbf{x}^*}
\newcommand{\bl}{\mathbf{x}^\prime}
\newcommand{\tabyes}{\textcolor{green}{\checkmark}}
\newcommand{\tabno}{\textcolor{red}{\texttimes}}
\usepackage{xurl}
\usepackage{authblk}
\hypersetup{colorlinks=true,linkcolor=blue!80,urlcolor=purple!80,citecolor=blue!80}
\title{Attributions Beyond\thanks{``Beyond'', because we do not focus on deep learning but optimization and operation research.} \ Neural Networks:\\ The Linear Program Case}

\author{\textbf{Florian Peter Busch}\textsuperscript{\rm 1,3,$\dagger$} \quad
\textbf{Matej Zečević}\textsuperscript{\rm 1}
\quad \textbf{Kristian Kersting}\textsuperscript{\rm 1-3} \quad \textbf{Devendra Singh Dhami}\textsuperscript{\rm 1,3}}
\affil{\textsuperscript{\rm 1}Computer Science Department, TU Darmstadt, \textsuperscript{\rm 2}Centre for Cognitive Science, TU Darmstadt,\\
\textsuperscript{\rm 3}Hessian Center for AI (hessian.AI), \textsuperscript{\rm $\dagger$}correspondence:\ \texttt{florian.busch@tu-darmstadt.de}\\
\vspace{-.5cm}
}
\author{%
  Florian Peter Busch \\
  Artificial Intelligence and Machine Learning Lab\\
  TU Darmstadt\\
  64289 Darmstadt \\
  \texttt{florian_peter.busch@tu-darmstadt.de} \\
}

\begin{document}

\maketitle

\begin{abstract}
    Linear Programs (LPs) have been one of the building blocks in machine learning and have championed recent strides in differentiable optimizers for learning systems. While there exist solvers for even high-dimensional LPs, understanding said high-dimensional solutions poses an orthogonal and unresolved problem. We introduce an approach where we consider neural encodings for LPs that justify the application of attribution methods from explainable artificial intelligence (XAI) designed for neural learning systems. The several encoding functions we propose take into account aspects such as feasibility of the decision space, the cost attached to each input, or the distance to special points of interest. We investigate the mathematical consequences of several XAI methods on said neural LP encodings. We empirically show that the attribution methods Saliency and LIME reveal indistinguishable results up to perturbation levels, and we propose the property of Directedness as the main discriminative criterion between Saliency and LIME on one hand, and a perturbation-based Feature Permutation approach on the other hand. Directedness indicates whether an attribution method gives feature attributions with respect to an increase of that feature. We further notice the baseline selection problem beyond the classical computer vision setting for Integrated Gradients.% Quote by reviewer: ``This paper is not very easy to follow. For example, in the abstract, terms like LIME, Saliency, Directedness appears without any interpretation. It's better to have a one-sentence interpretation before it first occurs.''
\end{abstract}

\section{Introduction: Explaining an LP}\label{sec:intro}
With the great rise in popularity of Deep Learning in recent years which was corroborated by its tremendous success in various applications \citep{krizhevsky2012imagenet, mnih2013playing,vaswani2017attention}, the popularity of methods which help to understand such models has increased as well \citep{sundararajan2017axiomatic,selvaraju2017grad,hesse2021fast}. The latter works constitute a new sub-field within artificial intelligence (AI) research often referred to as explainable AI (XAI). While XAI has tried a vast variety of methods and techniques to unravel the ``blackbox'' of deep learning models, many restrictions can be found when it comes to the notion of explainability or interpretability that is expected and sought \citep{stammer2021right}. To therefore move beyond simple ``heat-map'' type of attributions, explainable interactive learning research (XIL; see for instance Teso et al.\cite{teso2019explanatory}) poses one such alternative. In this work, however, we move ``beyond'' classical attribution in the sense that we consider alternate models instead of alternate data streams or alternate definitions. That is, we consider Linear Programs (LP).
While there has been considerable progress in increasing the understanding of neural networks (NNs), the complexity of deep models has received significant attention from the field of XAI, even though other fields might benefit from such techniques as well---for example LPs from mathematical optimization.

% \begin{figure}[t!]
%     \centering
%     \includegraphics[width=\linewidth]{images/Figure1.pdf}
%     \caption{\textbf{Schematized Overview.} For the first time, we apply XAI to LPs. To justify said application, we first propose an encoding $\phi$ of the initial LP which is then being used for learning a NN. Subsequently, said NN is analyzed using XAI. (Best viewed in color.)}
%     \label{fig:intro}
% \end{figure}

\begin{wrapfigure}[14]{R}{0.5\linewidth}
    \centering
    \includegraphics[width=\linewidth]{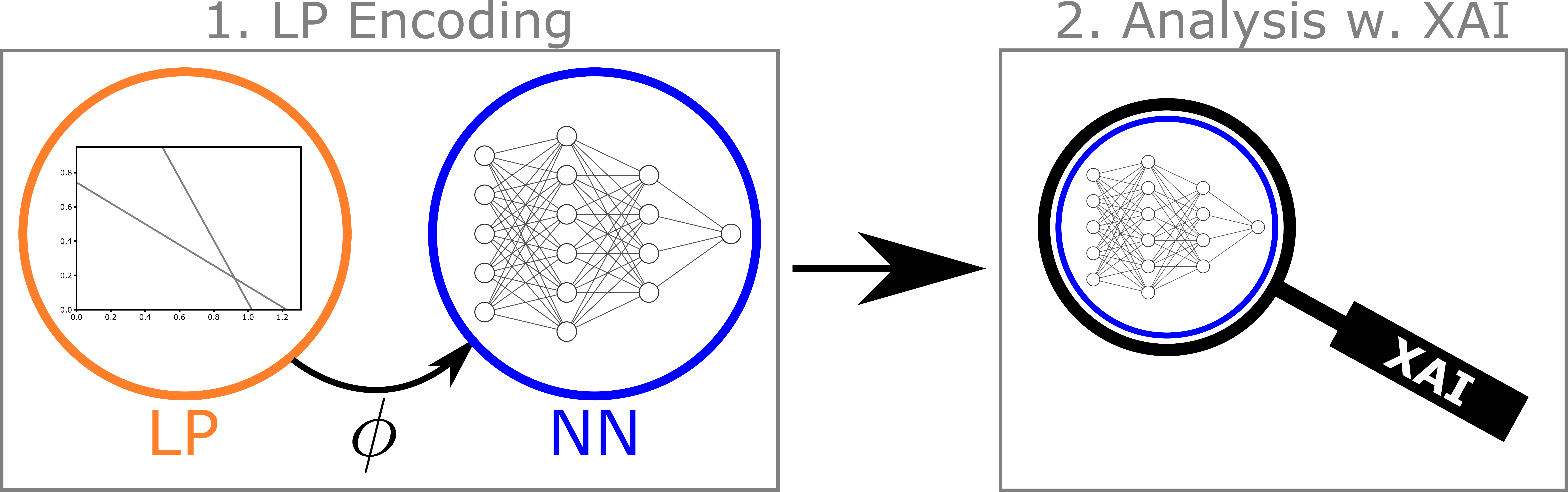}
    \caption{\textbf{Schematized overview.} For the first time, we apply XAI to LPs. To justify said application, we first propose an encoding $\phi$ of the initial LP which is then being used for learning a NN. Subsequently, said NN is analyzed using XAI. (Best viewed in color.)}
    \label{fig:intro}
\end{wrapfigure}

Interestingly, LPs are---at first glance---on the opposite spectrum of model complexity, they give the impression of being ``whitebox''.
Representing problems with linear constraints and relationships, LPs can be useful and in fact are used for a multitude of real world problems and there exist many solvers capable of solving even high-dimensional LPs.
It is even possible to use, solve, and backpropagate through LPs in a NN~\cite{paulus2021comboptnet,ferber2020mipaal}.
While such a solver can be used to calculate the optimal solution of a LP, some applications could benefit from an increased understandability of their underlying LPs. If LPs were to be ``properly explainable'', such that this refers to an intuitive, human level understanding, then this would have incredible implications for science and industry (well beyond AI research). For instance, energy systems researcher could design sustainable infrastructure to cover long-term energy demand \citep{schaber2012transmission}. Also, LP explainability naturally comes with strong implications within AI research, for instance in \emph{quantifying uncertainty and probabilistic reasoning} as in MAP inference \citep{weiss2012map}.

%To facilitate such understanding, we propose to combine the field of XAI and LPs in an approach which aims to utilize the methods from XAI in order to increase the understanding of LPs.
To facilitate such understanding, we propose to combine the field of XAI and LPs with the goal of utilizing the methods from XAI in order to increase the understanding of LPs.
Figure~\ref{fig:intro} illustrates our basic approach.
Given an LP, we use an encoding function $\phi$ to create data based on that LP which is then used to train a NN.
In the second step, we use common XAI methods mostly designed for use in Deep Learning to analyze the so trained NN.
Our work sets itself apart from previous work on sensitivity analysis~\cite{bazaraa2008linear} or infeasibility analysis~\cite{chinneck2007feasibility} by considering properties beyond feasibility and enabling the use of XAI methods on linear programs.

% For "bazaraa2008linear": Google scholar says "Sherali, Hanis D" instead of "Sherali, Hanif D", so I changed that manually, or should I have not done that?
% Quote by reviewer: ``Almost since the inception of linear programming, there has been a large body of work on on "explaining" linear programs, either focusing on sensitivity analysis or on the analysis of infeasibility. The paper overlooks this prior work.''

%Overall, we make the following contributions: (1) We introduce an encoding function which distinguishes feasible from infeasible instances, a function which includes the cost of LP solutions, two functions focusing on the constraints in particular, and another function using the vertices of the LP.
Overall, we make the following contributions: (1) We introduce an encoding function which distinguishes feasible from infeasible instances, a function which includes the cost of LP solutions, two functions focusing on the constraints, and a function using the LP vertices.
(2) We look at the attribution methods and compare them depending on these encodings. %For our experiment, we consider an easily visualizable 2-dimensional LP.
(3) We once again explain the importance of the selection of an appropriate baseline for IG.
(4) We show similarities between Saliency and LIME and we propose the property of \emph{Directedness} as the main discriminative criterion between Saliency and LIME on one hand, and our Feature Permutation approach on the other hand.

We make our code publicaly available at: {\footnotesize \url{https://cutt.ly/gHE0Nkx}}

\section{Background and Related Work}
% \todoR{@Dev: references.}

{\bf Optimization Using Linear Program.} A linear program (LP) consists of a cost vector $\mathbf{c} \in \mathbb{R}^n$ and a number of inequality constraints specified by $\mathbf{A} \in \mathbb{R}^{m \times n}$ and $\mathbf{b} \in \mathbb{R}^m$.
The objective of an LP is to find a point (or instance) $\mathbf{x} \in \mathbb{R}^n$ which minimizes the cost $\mathbf{c}^T\mathbf{x}$, is non-negative ($\mathbf{x} \geq \mathbf{0}$), and does not violate the constraints specified by $\mathbf{A}$ and $\mathbf{b}$ in the form of $\mathbf{A}\mathbf{x} \leq \mathbf{b}$. In short, given $\mathbf{c}, \mathbf{A}, \mathbf{b}$ find an instance $\mathbf{x}$ such that $\mathbf{c}^T\mathbf{x}$ is minimal, subject to $\mathbf{A}\mathbf{x}\leq \mathbf{b}$ and $\mathbf{x}\geq \mathbf{0}$.

% \begin{align}
% \begin{split}
%     &\min \mathbf{c}^T\mathbf{x}\\
%     \text{subject to}\quad&\mathbf{A}\mathbf{x}\leq \mathbf{b}\\
%     \text{and}\quad&\mathbf{x}\geq \mathbf{0}\text{.}
% \end{split}
% \end{align}
There can either be no solution, one solution or infinitely many solutions \citep{hoffman1953computational}. The minimization task is the objective function and can also be written as a maximization task of the negative cost (i.e. \emph{maximizing the gain} instead of \emph{minimizing the cost}).
%Depending on the application, we talk about the \emph{maximization of the gain} instead of the \emph{minimization of the cost}.
In this paper, unless stated otherwise, when talking about the \emph{constraints} we refer to $\mathbf{A}\mathbf{x} \leq \mathbf{b}$ and not $\mathbf{x} \geq \mathbf{0}$.
% Here we call $\mathbf{A}$ the \emph{constraint matrix} and $\mathbf{b}$ the \emph{constraint vector}.
When referring to a certain number of constraints in a specific LP, we refer to such a number of rows in $\mathbf{A}$ and corresponding elements $\mathbf{b}$.

{\bf Attribution Methods from XAI.} We follow the standard notion defined in \cite{sundararajan2017axiomatic}:
\begin{definition}
Let $F : \mathbb{R}^n \rightarrow \mathbb{R}$ be a NN and $\mathbf{x} = (x_1, \dots, x_n) \in \mathbb{R}^n$ denote an input. Then we call $A_F(\mathbf{x}) = (a_1, \dots, a_n) \in \mathbb{R}^n$ attribution where $a_i$ is the contribution of $x_i$ for prediction $F(\mathbf{x})$.
\end{definition}
Therefore, the attribution $\mathbf{A}_F(\mathbf{x})$ for the instance $\mathbf{x}$ consists of one attribution value for each feature of $\mathbf{x}$. Such an attribution value $\mathbf{a}_i$ for the $i$-th feature describes the contribution of that feature for the output. How exactly this contribution should be understood depends on the attribution method. %At least for all attribution methods considered in this paper, a contribution $\mathbf{a}_i$ should be $0$ if the corresponding feature $\mathbf{x}_i$ has no impact on the output at all.
% Quote by reviewer: ``Definition 1 should be stated more rigorously. What is the "contribution" of a variable to a prediction?'', other reviewer: ``In definition 1, it's not clear what are $\mathbf{A}_F(\mathbf{x})$ and their coordinate $\mathbf{a}_i$.''

The attribution methods here were chosen with the aim of enabling a comparison of different approaches. ``Captum'' was used to apply these methods \citep{kokhlikyan2020captum}. Subsequently, we briefly cover each of the attribution methods relevant in this works analysis.

{\it Integrated Gradients} \cite{sundararajan2017axiomatic} proposed an attribution method for deep networks which calculates attribution relative to a baseline. Formally, we are given $\text{IG}_i(\mathbf{x}) = (x_i - x^\prime_i) \times \int_{\alpha=0}^1 \frac{\partial F(\bl+\alpha \times (\mathbf{x}-\bl))}{\partial x_i} d\alpha$
% \begin{equation}
%     \text{IG}_i(\mathbf{x}) = (x_i - x^\prime_i) \times \int_{\alpha=0}^1 \frac{\partial F(\bl+\alpha \times (\mathbf{x}-\bl))}{\partial x_i} d\alpha
% \end{equation}
where $x_i$ is the $i^{th}$ element of $\mathbf{x}$ and $\bl$ is the baseline.

{\it Saliency} \cite{simonyan2014deep} proposed a straight forward approach in which the attribution is obtained by taking the predictive derivative with respect to the input. Therefore $\text{SAL}_i(\mathbf{x}) = \frac{dF}{dx_i}(\mathbf{x})$.
% \begin{equation}
%     \text{SAL}_i(\mathbf{x}) = \frac{dF}{dx_i}(\mathbf{x})
% \end{equation}

{\it Feature Permutation.} In the following we consider a perturbation-based notion to feature permutation, to allow for comparison in later sections. Feature Permutation \citep{breiman2001random} requires multiple instances for calculating the attribution.
The general idea is to use a batch of instances, iterate over every feature, permute the values of the respective feature in that batch, and then derive the attribution for each feature and instance by calculating the difference in predictive quality before and after the permutation. Since we want to have attributions for only single instances, we first define the \emph{feature importance} ($FI$) function as acting on batches, that is
$FI : \mathcal{X} \subset \mathbb{R}^{m \times n} \rightarrow \mathbb{R}^{m \times n}\text{,}$
where $\mathcal{X}$ is the space of all batches of $m$ instances with $n$ features. $FI$ takes a batch of instances and calculates attribution for each instance and feature by applying permutation.
In order to obtain attribution for a single instance, we generate instances around the input instance using perturbation.
We say that function $p(\mathbf{x}, m) \in \mathbb{R}^{(m+1) \times n}$
generates $m\in \mathbb{N}$ random perturbations of $\mathbf{x}$ and returns those perturbations and $\mathbf{x}$ ($\mathbf{x}$ in the last column).
With $FI$ and $p$, we can now apply the feature permutation algorithm of ``Captum'' on a single instance using $FI \circ p$ but we still need to decide on how we use this attribution for this batch to obtain an attribution for just one instance.
To this end, we generate one perturbed instance around the input point, calculate the attribution of this batch of two points using feature permutation $\mathbf{P} = (FI_i \circ p)(\mathbf{x}, 1)$ (where $FI_i$ returns the feature importances of feature $i$ only) and return only the attribution with respect to the original input instance.
In order to decrease the impact of randomness, this is done 10 times and the attributions are averaged.
Therefore, this feature permutation approach $\text{FP}$ is given by $\text{FP}_i(\mathbf{x}) = \frac{\sum_{j=1}^{10} \mathbf{P}_{:,2}}{10}\text{,}$
% \begin{equation}
%     \text{FP}_i(\mathbf{x}) = \frac{\sum_{j=1}^{10} \mathbf{P}_{:,2}}{10}\text{,}
% \end{equation}
where $\mathbf{P}_{:,2}$ is the second column of $\mathbf{P}$ (the column representing the attribution with respect to the original input instance).

{\it LIME} \cite{ribeiro2016should} used a strategy entirely different from the aforementioned attribution methods. The key idea of LIME is to train a surrogate model. While this model will only be able to reliably predict accurate results for the area around the input instance, this reduced complexity aims to make this additional model more \emph{interpretable}. We have $\text{LIME}_i(\mathbf{x}) = \mathbf{w}_{IM}$ where $\mathbf{w}_{IM}$ are the weights of the linear interpretable model surrounding $\mathbf{x}$. Then we simply state $\text{LIME}_i(\mathbf{x}) = R(p(\mathbf{x}), F_B(p(\mathbf{x})))$
% \begin{equation}
%      \text{LIME}_i(\mathbf{x}) = R(p(\mathbf{x}), F_B(p(\mathbf{x})))
% \end{equation}
where $p: \mathbb{R}^n \rightarrow \mathbb{R}^{j \times n}$ is a function which generates a batch of perturbed data $\mathbf{X_P} \in \mathbb{R}^{j \times n}$ , $F_B$ is a batch-version of $F$ ($F_B: \mathbb{R}^{j \times n} \rightarrow  \mathbb{R}^j$, $F_B(\mathbf{X}_P) = (F(x_1), \dots, F(x_j))$) and $R$ is a ridge regression model ($R(\mathbf{X}_P, F_B(\mathbf{X}_P)) = \text{argmin}_w ||F_B(\mathbf{X}_P)-\mathbf{X}_P w||^2_2 + ||w||^2_2$).

{\bf General Properties of Attribution Methods.} The are several well known properties of attribution methods which will be useful later.
An attribution method is {\it (a) Gradient Based, } if the attribution method relies on calculating gradients \citep{ancona2019gradient}.
It is {\it (b) Perturbation Based,} if the attribution method uses perturbations to generate data around the input point \citep{zeiler2014visualizing}.
{\it (c) Completeness,} is satisfied if the attribution method relies on a baseline $\bl$, and $\sum_{i=1}^n A_F(\mathbf{x}) = F(\mathbf{x}) - F(\bl)$ is true~\citep{sundararajan2017axiomatic}.
And we refer to {\it (d) Randomness,} if randomness is involved in the calculation for the attribution.
The impact of randomness can be reduced at cost of calculation time by increasing number of samples, steps, etc.

\section{Encoding Priors for Linear Programs}
\label{sec:lp_enc}
In recent times, XAI methods have come to be primarily focused on neural methodologies \citep{gunning2019darpa}. To justify the usage of XAI for LPs, we consider their similarities with neural models. Therefore, we first will consider how to encode LPs in a ``neural'' manner. One of the most obvious approaches here could be to use $\mathbf{c}$, $\mathbf{A}$, and $\mathbf{b}$ as inputs for the NN and the optimal solution $\sol$ for the output.
This approach would give us attribution for the inputs, so $\mathbf{c}$, $\mathbf{A}$, and $\mathbf{b}$.
While we believe that such attributions would also carry valuable information, in this paper we focus on the information we can obtain from looking at one specific, single LP, so we consider $\mathbf{c}$, $\mathbf{A}$, and $\mathbf{b}$ fixed.

% For this purpose, we need to think of which information we can get and learn considering a specific LP.
If we only train with a constant $\mathbf{c}$, $\mathbf{A}$, and $\mathbf{b}$ as inputs, the NN will simply learn to output the optimal solution $\sol$ without actual ``learning''.
Therefore, we need a task from which we can construct a dataset with different inputs and corresponding outputs.
We do so by inputting instances from the LP ($\mathbf{x} \in \mathbb{R}^n$).
Now, we can define what the output should be and thus the corresponding learning problem.
We refer to such a problem as an \emph{encoding} ($\phi$) of the original LP.
Note, that this approach requires a new NN to be trained for each encoding and LP, thereby learning the relevant properties of the respective encoding well enough so that XAI methods can be applied correctly.
Obtaining enough useful data and finding good NN hyperparameters can be challenging.
Following paragraphs cover various reasonable encodings that we initially propose and then systematically investigate.

\paragraph{Feasibility Encoding.} In a straight forward manner, we can simply distinguish between \emph{feasible} instances, i.e. instances which do not violate any constraints, and \emph{infeasible} instances, i.e. instances which violate at least one constraint.
Using binary coding, we have $\phi(\mathbf{x}) =
        \begin{cases}
            1 & (\mathbf{A}\mathbf{x} \leq \mathbf{b}) \\
            0 & (\mathbf{A}\mathbf{x} > \mathbf{b}).
        \end{cases}$.
% \begin{equation}
%     \phi(\mathbf{x}) =
%         \begin{cases}
%             1 & (\mathbf{A}\mathbf{x} \leq \mathbf{b}) \\
%             0 & (\mathbf{A}\mathbf{x} > \mathbf{b}).
%         \end{cases}
% \end{equation}
Note that for any LP encoding we will keep the LP constraints $\mathbf{A},\mathbf{b}$ constant, therefore, in our notion the encoding function only depends on the optimization variable $\mathbf{x}$.

\paragraph{Gain–Penalty Encoding.}
In this approach, the feasible instances get assigned their corresponding gain and the score of an infeasible instance now depends on how much it violates the constraints.
This is done by finding an $\epsilon$-close (ideally the closest) feasible point, calculating the gain of that point and reducing it, depending on the distance between these two points.
As a consequence, infeasible instances with only a small constraint violation have almost the same gain as the closest feasible instances but infeasible instances with large violations get assigned increasingly smaller scores (gains).
To some degree, real world applications might justify this approach since the constraints are not completely prohibitive but rather that violations of them are simply too costly to be efficient.
The Gain–Penalty encoding is defined as $\phi(\mathbf{x}) =
        \begin{cases}
            \mathbf{c}^T\mathbf{x} & (\mathbf{A}\mathbf{x} \leq \mathbf{b}) \\
            \mathbf{c}^T\mathbf{x}_f (1- \min(1, \frac{||\mathbf{x}-\mathbf{x}_f||_2}{||\mathbf{x}_f||_2})) & (\mathbf{A}\mathbf{x} > \mathbf{b})\text{,}
        \end{cases}$,
% \begin{equation}\label{eq:gainpen}
%     \phi(\mathbf{x}) =
%         \begin{cases}
%             \mathbf{c}^T\mathbf{x} & (\mathbf{A}\mathbf{x} \leq \mathbf{b}) \\
%             \mathbf{c}^T\mathbf{x}_f (1- \min(1, \frac{||\mathbf{x}-\mathbf{x}_f||_2}{||\mathbf{x}_f||_2})) & (\mathbf{A}\mathbf{x} > \mathbf{b})\text{,}
%         \end{cases}
% \end{equation}
where $\mathbf{x}_f {=} \text{argmin}_{\mathbf{x}_i} ||\mathbf{x}_i - \mathbf{x}||_2$ is the closest feasible instance in the setting where $\mathbf{c}, \mathbf{A}, \mathbf{b}, > \mathbf{0}$. Naturally, there exist various, sensible variations to this formulation.

\paragraph{Boundary Distance Encoding.}
For this encoding, we investigate the boundary between feasible and infeasible instances.
By calculating the minimum of $\mathbf{b}-\mathbf{A}\mathbf{x}$, we can find out if an instance lies on the boundary and also obtain a score indicating how large the biggest violation is or, for feasible instances, how much space there is until the closest constraint would be violated: $\phi(\mathbf{x}) = \min(\mathbf{b}-\mathbf{A}\mathbf{x})$.
% \begin{equation}
%     \phi(\mathbf{x}) = \min(\mathbf{b}-\mathbf{A}\mathbf{x})\text{.}
% \end{equation}
We can also take the absolute of this encoding $|\phi(\mathbf{x})|$ (Absolute Boundary Distance encoding) such that we can treat the output as a distance to the boundary (\emph{margin}), changing attribution method behavior.

\paragraph{Vertex Distance Encoding.}
Here, the distance to the nearest vertex is used. A vertex of an LP is any feasible point which lies on either the intersection of two constraints or on the intersection of a constraint with an axis (or at the origin).
Note, that any optimal solution has to lie on one of those vertices but since this approach does not include the cost vector $\mathbf{c}$, it does not contain any information about which of those vertices is an optimal solution.
Overall, the Vertex Distance encoding is defined as $\phi(\mathbf{x}) = \min_{\mathbf{x}} \mathcal{V}$,
% \begin{equation}
%     \phi(\mathbf{x}) = \min_{\mathbf{x}} \mathcal{V},
% \end{equation}
with $\mathcal{V}=\{||\mathbf{x}-\mathbf{x}_v||_2 | \mathbf{x}_v \in \mathbf{X}_v\}$
% \begin{equation}
%     \mathcal{V}=\{||\mathbf{x}-\mathbf{x}_v||_2 | \mathbf{x}_v \in \mathbf{X}_v\}
% \end{equation}
where $\mathbf{X}_v$ denotes the set of vertices.

\section{Properties of LP Encodings and Attribution Methods}
Upon establishing various, sensible encodings $\phi$, we now move on to the proposal of reasonable key properties for the discussed $\phi$ that we can come back to when we discuss how attribution methods behave on different encodings.
To ease notation later on, we define the set $\mathcal{X}\subset \mathbb{R}^n_{\geq 0}$ such that the properties are only related to instances which are not infeasible for every single LP by construction (since $\mathbf{x} \geq \mathbf{0})$. We propose the following key properties.

\paragraph{Continuity.}
We call an LP encoding $\phi$ continuous if $\forall \mathbf{x}, \mathbf{k} \in \mathcal{X}  \ldotp (\lim_{\mathbf{x} \rightarrow \mathbf{k}} \phi(\mathbf{x}) = \phi(\mathbf{k})).$
% $$\forall \mathbf{x}, \mathbf{k} \in \mathcal{X}  \ldotp (\lim_{\mathbf{x} \rightarrow \mathbf{k}} \phi(\mathbf{x}) = \phi(\mathbf{k})).$$

\paragraph{Distinguish Class/Distinguish Boundary.}
% If an LP encoding $\phi$ computes values in such a way that you can infer from the output whether the point is feasible or infeasible, we say that $\phi$ satisfies Distinguish Class.
If an LP encoding $\phi$ computes values in such a way that feasibility of the point can be inferred from its output, we say that $\phi$ satisfies Distinguish Class.
Formally,
$ \exists f, \forall \mathbf{x} \in \mathcal{X} \ldotp(\mathbf{A}\mathbf{x} \leq \mathbf{b} \leftrightarrow f(\phi(\mathbf{x}))).$
In a similar way, Distinguish Boundary is satisfied if we can infer whether a point lies on the decision boundary:
$ \exists g, \forall \mathbf{x} \in \mathcal{X} \ldotp(\mathbf{A}\mathbf{x} = \mathbf{b} \leftrightarrow g(\phi(\mathbf{x}))).$

\paragraph{Boundary Extrema.}
This property is concerned with whether the boundaries of the LP are at the extrema of $\phi$.
There are multiple possibilities to define such a property.
We postulate that a method satisfies Boundary Extrema if an extremum (maximum or minimum) of $\phi$ lies on the decision boundary and if this extremum does not appear outside the boundary.
The extremum may appear more than once on the boundary itself.
% Let $P_i:=\mathbf{A}\mathbf{x}_i-\mathbf{b} = \mathbf{0}$, then formally we have that there $\exists \mathbf{x}_i,\forall\mathbf{x}_j \in \mathcal{X}$ such that 
% \begin{align*}
% P_i \land ((\phi(\mathbf{x}_i) > \phi(\mathbf{x}_j) \lor P_j) \lor (\phi(\mathbf{x}_i) < \phi(\mathbf{x}_j) \lor P_j)).
% \end{align*}
Let $P_i:=\min(\mathbf{A}\mathbf{x}_i-\mathbf{b}) = \mathbf{0}$, then formally we have that there $\exists \mathbf{x}_i,\forall\mathbf{x}_j, \mathbf{x}_k \in \mathcal{X}$ such that $P_i \land ((\phi(\mathbf{x}_i) > \phi(\mathbf{x}_j) \lor P_j) \lor (\phi(\mathbf{x}_i) < \phi(\mathbf{x}_k) \lor P_k)).$
% \begin{align*}
% P_i \land ((\phi(\mathbf{x}_i) > \phi(\mathbf{x}_j) \lor P_j) \lor (\phi(\mathbf{x}_i) < \phi(\mathbf{x}_k) \lor P_k)).
% \end{align*}

%Table~\ref{tab:LP_enc_prop} summarizes which of our proposed encoding methods satisfy which of the properties presented here.

% \begin{table}[ht]
%     \centering
%     \caption{\textbf{Encodings and their Properties.} The abbreviations are: F~=~Feasibility, GP~=~Gain–Penalty, BD~=~Boundary Distance, ABD~=~Absolute Boundary Distance, VC~=~Vertex Distance.}\label{tab:LP_enc_prop}
%     \begin{tabular}{l|c|c|c|c|c}
%         & F & GP & BD & ABD & VD   \\\hline
%         Continuity & \tabno & \tabyes & \tabyes & \tabyes & \tabyes \\
%         DistinguishClass & \tabyes & \tabno & \tabyes & \tabno & \tabno \\
%         DistinguishBoundary & \tabno & \tabno & \tabyes & \tabyes & \tabno \\
%         BoundaryExtrema & \tabno & \tabyes & \tabno & \tabyes & \tabyes \\
%     \end{tabular}
% \end{table}

\begin{table}[t]
    \centering
    \caption{\textbf{Properties of encodings and attribution methods.} On the left: properties of encodings, on the right: properties of attribution methods. The abbreviations are: F~=~Feasibility, G~=~Gain–Penalty, B~=~Boundary Distance, A~=~Absolute Boundary Distance, V~=~Vertex Distance, S~=~SAL, L~=~LIME. Regarding Neighborhoodness (*): Not binary, thus requires alternate interpretation. We argue that IG generally has the least Neighborhoodness because the path from baseline to input can be very long. Saliency always considers a smaller area than the perturbation-based approaches FP and LIME.}\label{tab:properties}
    \begin{tabular}[t]{l|c|c|c|c|c}
        & F & G & B & A & V   \\\hline
        Continuity & \tabno & \tabyes & \tabyes & \tabyes & \tabyes \\
        DistinguishClass & \tabyes & \tabno & \tabyes & \tabno & \tabno \\
        DistinguishBoundary & \tabno & \tabno & \tabyes & \tabyes & \tabno \\
        BoundaryExtrema & \tabno & \tabyes & \tabno & \tabyes & \tabyes \\
    \end{tabular}
    \begin{tabular}[t]{l|c|c|c|c}
        & IG & S & FP & L  \\\hline
        GradientBased & \tabyes & \tabyes & \tabno & \tabno \\
        PerturbationBased & \tabno & \tabno & \tabyes & \tabyes \\
        Completeness & \tabyes & \tabno & \tabno & \tabno \\
        Randomness & \tabno & \tabno & \tabyes & \tabyes \\
        Neighborhoodness (*) & \tabno & \tabyes \tabyes & \tabyes & \tabyes \\
        Directedness & \tabno & \tabyes & \tabno & \tabyes \\
    \end{tabular}
\end{table}

% \section{Revisiting Properties of Attribution Methods}
% In addition to well known properties of attribution methods, we now propose a new concept as well as another property for attribution methods which will be useful for discussing differences of those methods afterwards.
Previously, we stated some well known properties of attribution methods.
In addition to those, we now propose a new concept as well as another property for attribution methods which will be useful for discussing differences of attribution methods afterwards.

{\bf Neighborhoodness.} Being more of a concept rather than a property, Neighborhoodness describes how large a region around the input point is considered for the attribution. For example, Neighborhoodness of perturbation-based approaches usually depends on the perturbation's ``strength''.

{\bf Directedness.} We say that Directedness is satisfied if the attribution for a feature indicates how increasing that feature would change the output.
The direction here is important, so the attribution is always considered with an increase of that feature in mind (for example, attribution would be negative if increasing that feature decreases the output).
This increase might be local or over a larger interval, therefore the property does not directly depend on the Neighborhoodness of the attribution method.

We give an overview of the previously introduced properties of our LP encodings and on how the considered attribution methods fall into properties such as Neighborhoodness, Directedness, and other established properties in Table~\ref{tab:properties}.

% \begin{table}[ht]
%     \centering
%     \caption{\textbf{Attribution Methods and their Properties.} Regarding Neighborhoodness (*): Not binary, thus requires alternate interpretation. We argue that IG generally has the least Neighborhoodness because the path from baseline to input can be very long. Saliency always considers a smaller area than the perturbation-based approaches FP and LIME.}\label{tab:LP_attr_meth_prop}
%     \begin{tabular}{l|c|c|c|c}
%         & IG & SAL & FP & LIME  \\\hline
%         GradientBased & \tabyes & \tabyes & \tabno & \tabno \\
%         PerturbationBased & \tabno & \tabno & \tabyes & \tabyes \\
%         Completeness & \tabyes & \tabno & \tabno & \tabno \\
%         Randomness & \tabno & \tabno & \tabyes & \tabyes \\
%         Neighborhoodness (*) & \tabno & \tabyes \tabyes & \tabyes & \tabyes \\
%         Directedness & \tabno & \tabyes & \tabno & \tabyes \\
%     \end{tabular}
% \end{table}

\section{Empirical Illustration}
\label{sec:emp}
% : Extended Discussion for Example LP}
We analyze the encodings, the properties, and the implications from the previous three sections empirically using an illustrative LP setting.

{\bf Experimental Setup.} To compare different encodings, we consider an LP that allows for 2d-visualization of its polytope. Further, we use a NN with 7 layers to train a model for each encoding introduced in Section~\ref{sec:lp_enc}.
Another experiment only trains a NN for the feasibility encoding but uses an LP with 5 dimensions.
For all experiments, we generate 100,000 random instances such that the complete LP polytope (set of feasible solutions) is covered and that about as many feasible and infeasible instances are being generated.
For integrated gradients we use the baseline $\bl = \mathbf{0}$.
In our experiments, FP and LIME use perturbations by generating instances around the input point, up to a set maximum perturbation.
We use a Ridge Regression model with 2 dimensions for LIME.

Through our extensive empirical evaluations we aim to answer the following research questions:
[\textbf{Q\emph{i}.}] How do encodings change $X$'s attributions? (where \textbf{\emph{i}}${\in}\{1,2,3,4\}$ with $X$ being IG, SAL, FP and LIME respectively). [\textbf{Q5.}] How does the feasibility encoding perform on a higher dimensional example? [\textbf{Q6.}] How do all the discussed attributions (attribution methods) relate to each other?

\begin{figure*}
    \centering
    \includegraphics[width=\linewidth,origin=c]{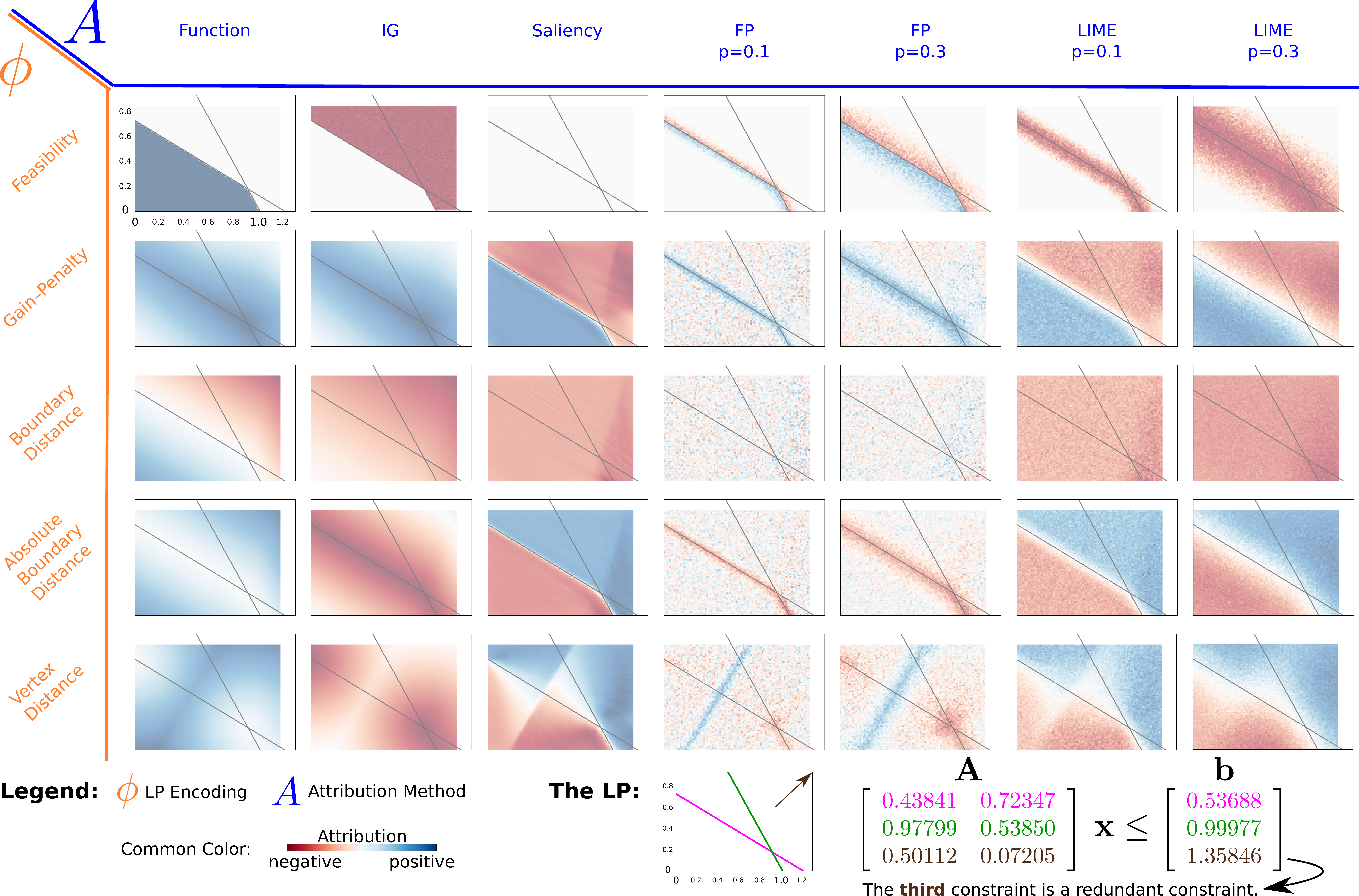}
    % \vspace{.25cm}
    \caption{\textbf{Overview-matrix of LP encodings and attribution methods.}
    All plots show a single LP with two features, one horizontal and one vertical, where the grey lines indicate the constraints for that LP. This matrix of plots shows the summed up attribution of both features (attribution sum). The encodings (orange, rows) are plotted against the different attribution methods (blue, columns).
    % The bottom part (enclosed in the grey box titled ``Feature Attributions'') shows attribution for both features on the Vertex Distance encoding as an example.
    For FP and LIME, $\text{p}$ indicates the maximum possible perturbation in any direction. The (rounded) numbers for the constraints of that LP are shown on the bottom right.
    %The constraint of the third row in $\mathbf{A}$ is a redundant constraint, it does not restrict the set of feasible instances.
    (Best viewed in color.)
    }
    \label{fig:results}
\end{figure*}

\paragraph{Overview of Results.}
% In Fig.\ref{fig:results} we show the encodings versus attribution methods matrix (further revealing the fine-grained horizontal/vertical feature attributions as an illustrative example for the Vertex Distance encoding).
In Fig.\ref{fig:results} we show the encodings versus attribution methods matrix (here only with the summed up feature attributions, refer to Supplementary for an example showing single feature attributions on the same LP).
We approximate the regions we generated data for with $100 \times 73$ pixels.
Note, that we ignored the vertex $(0, 0)$ for the Vertex Distance encoding by including our prior knowledge that it cannot be the optimal solution.
We refer to $\sum_{i=1}^n A_F(\mathbf{x})$ as the attribution sum.

\textbf{Q1. Integrated Gradients.}
The attribution sum for Integrated Gradients follows from its Completeness property: $\sum_{i=1}^n \text{IG}_i(\mathbf{x}) = \phi(\mathbf{x}) - \phi(\bl)\text{.}$
% $$\sum_{i=1}^n \text{IG}_i(\mathbf{x}) = \phi(\mathbf{x}) - \phi(\bl)\text{.}$$
Therefore, if Continuity, Distinguish Class, Distinguish Boundary, or Boundary Extrema is satisfied on $\phi$, that same property also applies to $\text{IG}$.
A benefit of Completeness is that the attribution for each feature indicates how large its contribution relative to the change in output compared to the baseline output is.
In contrast to other methods with no such property, this makes it so that every attribution has a tangible meaning and that the attribution for a single instance can be understood without other instances as a comparison.\footnote{Arguably, even IG can not explain instances individually, as every explanation requires a baseline reference.}

Of course, if Completeness is satisfied, the important part of any attribution is how the contribution is divided amongst the individual features.
% For example, consider the feature attribution for the Vertex Distance encoding in Figure~\ref{fig:results}.
For example, consider the Vertex Distance encoding in Figure~\ref{fig:results}.
Our baseline lies at $(0, 0)$, a point with a relatively large distance to any of the three vertices.
% The attribution sum is the smallest on the vertices because those have the smallest values.
% We can see that attribution for the top left vertex mainly comes from the vertical feature and attribution for the two right vertices from the horizontal feature.
% This is because the change of that respective feature is much more important for the change in distance to the respective vertex.
Here, the vertex on the bottom right only gets attribution for the horizontal feature, the vertex on the top left only attribution for the vertical feature as these respective features differ from the baseline (see this visualized in Supplementary).
This would be  different if a baseline on the top right was chosen.
Therefore, the main challenge when applying IG is choosing a sensible baseline.
% The main challenge when applying IG is choosing a sensible baseline.
Coming back to our previous example, maybe using a vertex as the baseline could carry more meaning, however then you would still have to decide which vertex to choose as this also influences the resulting attributions.

Overall, applying IG on LP encodings can result in useful and understandable attributions but those always must be interpreted with respect to the respective baseline.
In addition, choosing an appropriate baseline is often not obvious.

\textbf{Q2. Saliency.}
Since Salience simply calculates the local gradient, its general behavior is easily explained.
In accordance with satisfying Directedness, attribution for a feature is positive if the feature impact on the respective point is positive, negative if it is negative, and 0 otherwise.
Notably, this also means that there is no attribution on local extrema (see encodings satisfying Boundary Extrema in Figure~\ref{fig:results}).
Since Saliency has a very small Neighborhoodness, attribution usually requires a certain amount of additional information to be useful.
For example, attributions on and around a local maximum have different values, ranging from positive attribution where the value increases towards the maximum, to 0 exactly on the maximum, and to negative attribution afterwards.
All those attributions are ``correct'' but if you were to consider only one of those points and have no knowledge about the maximum there, you might draw false conclusions about the feature impact in that area.
In this example, prior knowledge about that local maximum or at least considering multiple other points reduces that risk.

One interaction worth discussion is how Saliency behaves on points where the encoding function is not differentiable.
We can see this for the Absolute Boundary Distance encoding in Figure~\ref{fig:results}.
Due to the absolute value function, points on the boundary are not differentiable.
Still, the NN approximates this function in a differentiable way.
Not only does this result in a 0 attribution (gradient) exactly on the boundary but there is also a small area around the boundary where the gradient quickly, but not instantly, changes from 0 to the true gradient of the area around.
Other not differentiable points appear whenever an encoding does not satisfy Continuity.
For the Feasibility encoding, this results in an extreme attribution very close to the boundary, where the NN approximates this discontinuity with a very steep function.

To sum up, attribution here reflects very local changes w.r.t. an increase of input features.
Approximations of non-differentiable points by the NN influence Saliency attributions near those points.

\textbf{Q3. Feature Permutation.}
Due to its perturbations, FP focuses on the change of the output around the input point.
Since it does not satisfy Directedness, attribution in areas with a steadily changing output in one direction can average to 0, as the positive change in one and the negative change in the opposite direction cancel each other out.
If that change in both directions is equal, the only remaining attribution is a result of the Randomness in the perturbation process.

Even though Directedness is not satisfied, there is a difference between negative and positive attribution but it has to be understood differently.
Because FP considers the area around it without consideration of direction, local minima (maxima) can be observed to have negative (positive) attribution because permuting features of instances around it increases (decreases) their function output.\footnote{An increase of output leads to a negative attribution because the feature importance is calculated by the feature permutation algorithm as the original output minus the new output (see \url{https://captum.ai/api/feature_permutation.html}).} The degree of perturbation can be seen as a trade-off between precise, local results on the one hand and robust and emphasized attribution for larger areas on the other hand. All in all, attributions given by Feature Permutation consider any direction equally, thereby resulting in a focus on local minima and maxima of the encoded function.

\textbf{Q4. LIME.}
In Figure~\ref{fig:results}, we can now observe how LIME with smaller perturbation (Neighborhoodness) look increasingly similar to Saliency.
This is just as much true for the attributions of the features alone which, as described by Directedness, indicate how a feature impacts the output with respect to an increase of that feature.
The major difference when comparing the results of Saliency and LIME is that we can choose how large of an area around the input point should be considered by LIME.
So while small perturbations behave like Saliency, larger perturbations can help to focus on the more overarching characteristics and also make the procedure more robust against local errors and bad approximations of the NN.
On the other hand, small, local perturbations allow for a more precise attribution of the respective points.
Based on our experiments, the nature of LPs, and our noiseless data generation, we argue that usual downsides of very local perturbations might not be as significant.

On discontinuities, LIME, just like any method with a certain Neighborhoodness, can have very large attributions.
Here, the attribution, so the slope of the linear model, is steeper the smaller the perturbations.
Overall, LIME attributions show the same characteristics as those generated by Saliency.
However, LIME allows for changing the Neighborhoodness of the attribution, making it possible to both consider only very local changes or include a larger area around the point of interest.

\begin{figure}
    \centering
    \includegraphics[width=\linewidth]{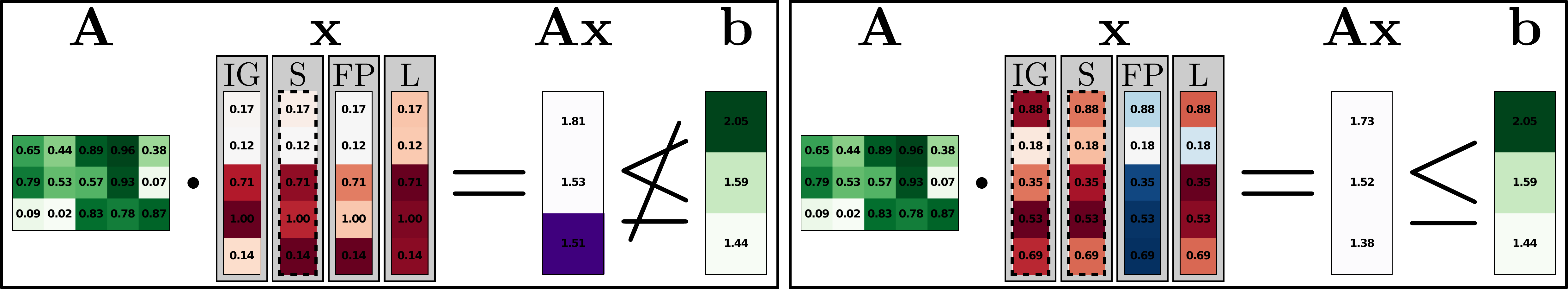}  % _Figure5_v2.pdf for the same figure but one below the other
    \caption{\textbf{5-dimensional feasibility experiment.} Here, an infeasible instance on the left and a feasible instance on the right are shown. In $\mathbf{A}$ and $\mathbf{b}$, green indicates the element size. Constraint violations are marked purple in $\mathbf{A}\mathbf{x}$. Attributions go from red (negative) to blue (positive). Attributions with dashed border have very small absolute attribution values ($<0.01$). All values are rounded.
    }
    \label{fig:fig_5dim}
\end{figure}

\textbf{Q5. Higher dimensional Experiment.}
To show that we can handle higher dimensions we consider an input $\mathbf{x}$ consisting of five dimensions and three constraints.
Figure~\ref{fig:fig_5dim} shows one infeasible and one feasible instance with the corresponding constraints and constraint violations.
Attributions are shown for all four attribution methods, with an individual color scale for each method and input.
As before, \emph{red} indicates negative attribution, \emph{white} zero attribution, and \emph{blue} positive attribution.
% Both Saliency attributions and one IG attribution are marked with dotted lines indicating that their attribution values are all very close to 0 (all smaller than $0.01$). 
The left instance in Figure~\ref{fig:fig_5dim} is infeasible as it violates the third constraint ($1.51 \nleq 1.44$).
All attribution methods here focus on the last three features which have the highest impact on the violation of that third constraint. %(the other two feature are multiplied with small values).
% SENTENCE FOR THAT:? In comparison to FP, LIME (L) also assigns some attribution to those because it considers multiple features which makes it possible for those to change the output class.
% In comparison to the IG baseline (zero vector), this changes the class and assigns negative attribution to features which can be seen focuses on the violated constraint, i.e. the third row in $\mathbf{A}$.
% For FP (so if only a single feature was changed by not more than $0.1$), also only the three features which play a significant role for the third constraint get negative attribution.
% That LIME also has some attribution for the first two feature can be explained by the fact that LIME considers multiple changes of features by up to $0.1$ which makes it possible for the first two features to increase enough to violate the second constraint while the third constraint is not violated anymore. % FP can only violate the second constraint but not at the same time "unviolate" the third, therefore changing the first feature never changes the output and has not attribution
For the feasible instance (right), the second feature has the least attribution (remember that FP has a different sign in such a scenario).
Here, both the second and the third constraint are somewhat close to violation and while the last three features are important for the last constraint, the first feature is important for the second constraint.
%Looking at the feasible instances for FP and LIME, it can be seen that those same last three features get attribution because increasing those could violate the third constraint (remember, that the FP approach uses a different sign in that case) but here the first feature is also more present as increasing that could violate the second constraint. % we can see a small model error here in the blue (positive) attribution of LIME in the second feature
The Saliency (S) attributions and one IG attribution are marked with dotted lines indicating small attribution values (under $0.01$).
This is the result of the same type of behavior previously mentioned in Q2.
The discontinuous decision boundary is approximated by the NN in a continuous manner, resulting in instances close to the boundary to have outputs close to, but not exactly $0$ or $1$.
This effect can even be useful, as the ratio of the feature attributions still contains information (without it, these attributions would all be $0$).%\footnote{IG, because the input prediction is equal to the baseline prediction and Saliency, because the local gradient is, in theory, 0.}.
% In theory, these attributions should all be $0$ because the output does not differ from the baseline label (IG) and because there is no change in output (gradient) at exactly these points (Saliency).
% In practice, however, the discontinuity at the decision boundary is approximated by the NN is a continuous manner with continuous outputs.
% Therefore, instances close to the boundary have a predicted output only close to, but not exactly $0$ or $1$.
% This effect can even be useful, as the ratio of the feature attributions still contains information (instead of all attributions being 0).
% A small threshold could also be applied if that behavior was undesired (then only instances very close to the boundary would get this "false/misleading" attribution).

\textbf{Q6. Relationships between Attribution Methods.}
Having examined the general behavior of the four attribution methods considered in this paper, we now go into detail about how they relate to each other.
First of all, we argue that IG differs from the other methods significantly.
While the ``information'' that IG uses to calculate its attribution consists of the path between baseline and input, all other three methods base their attribution on the area around the input.
Looking at our results in Figure~\ref{fig:results}, we can see that especially Saliency and LIME show noticeable similarities, particularly if smaller perturbations are used for LIME.

% \begin{figure}
%     \centering
%     \includegraphics[width=\linewidth]{images/Figure3.pdf}
%     \caption{\textbf{``LIME $\equiv$ Saliency.''} With increasingly smaller perturbations, LIME comes visibly closer to Saliency. The decrease in attribution scores for smaller perturbation is due to the regularizer of the Ridge Regression model.
%     }
%     \label{fig:fig3}
% \end{figure}
% \begin{figure}[t]
%     \centering
%     \begin{minipage}[b]{0.48\textwidth}
%     \centering
%     \includegraphics[width=\linewidth]{images/Figure3_v2.pdf}
%     \caption{\textbf{``LIME $\equiv$ Saliency.''} With increasingly smaller perturbations, LIME comes visibly closer to Saliency. The decrease in attribution scores for smaller perturbation is due to the regularizer of the Ridge Regression model.
%     }
%     \label{fig:fig3}
%     \end{minipage}%
%     \hfill
%     \begin{minipage}[b]{0.48\textwidth}
%     \centering
%     \includegraphics[width=\linewidth]{images/Figure4_v2.pdf}
%     \caption{\textbf{$\text{FP}{\rightarrow}\text{LIME}$.} If we make it so that FP distinguishes between forward and backward perturbations and scales attributions according to their distance from the input point, we can get results similar or even identical to LIME.}
%     \label{fig:fig4}
%     \end{minipage}%
% \end{figure}
\begin{figure}
    \centering
    \includegraphics[width=\linewidth]{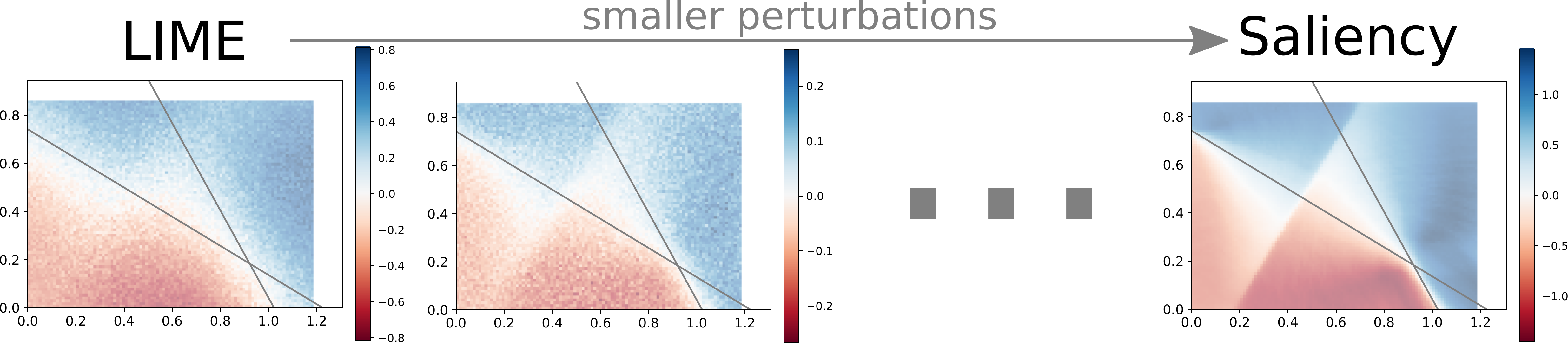}
    \caption{\textbf{``LIME $\equiv$ Saliency.''} With increasingly smaller perturbations, LIME comes visibly closer to Saliency. The decrease in attribution scores for smaller perturbation is due to the regularizer of the Ridge Regression model.
    }
    \label{fig:fig3}
\end{figure}
\vspace{1cm}
\textbf{Similarity of Saliency and LIME.}
The high-level description of how LIME functions is arguably similar to the underlying functioning of Saliency.
Could this intuitive similarity on a high-level hint towards more, possibly an equivalence under certain circumstances? 
\begin{conjecture}
For a perturbation $\longrightarrow$ zero, the attributions of LIME $\longrightarrow$ Saliency.
\end{conjecture}
Both approaches calculate a value for a specific point with respect to how the output values around this point change, in other words, they both satisfy Directedness.
So if we would perturb instances for LIME in such a way that they are infinitely close to the original input but not on the input itself (in which case we would simply have no attribution), then LIME approaches Saliency in the limit. We support this conjecture empirically (Fig.\ref{fig:fig3}).
From that perspective, the main difference between LIME and Saliency is that former allows for considering a larger area (perturbation function as a hyperparameter).
But even though attribution computation is different, we know from our experiments that FP and LIME use the same perturbations.
So if both those methods use the same ``information'' to calculate attributions, how can FP look different?

\textbf{How Feature Permutation differs from Saliency and LIME.}
Just like Saliency and LIME, FP uses nearby regions to determine attributions.
Since all those three methods appear to base their attribution on similar information, could we again postulate a result on the triangular relationship? Again, we conjecture on the approaches's similarity:
\begin{conjecture}
The main difference between Saliency and LIME on the one hand and FP on the other hand is Directedness. If you were to \emph{``insert''} Directedness into FP, you would get an approach which behaves almost identically to LIME and, thereby, also to Saliency (for small perturbations).
\end{conjecture}
Usually FP does not consider ``how'', i.e. in which direction a change happened, but only that it happened.
% For example, on a maximum, both Saliency and LIME (with small enough perturbations) would return 0 attribution because there is the same amount of increase leading up to it as decrease going away from it.\footnote{In both cases, the increase first and decrease later is with respect to an increase of the feature value, i.e. according to Directedness.}
For example, on a maximum, both Saliency and LIME (with small enough perturbations) would return little or no attribution because increase leading up to the maximum is canceled out by decrease going away from it.\footnote{In both cases, the increase first and decrease later is w.r.t.\ an increase of the feature value i.e., Directedness.}
However, FP only considers that there is an average decrease in any direction if a feature is changed, so if that decrease is large, then that feature must be important.
You can see that FP is not necessarily worse, you might very well argue that it should be preferred in this example.
But if FP would consider that \emph{decreasing} the feature \emph{decreases} the output and \emph{increasing} the feature also \emph{decreases} the output it could use this information of direction (Directedness) to also return an attribution akin to Saliency and LIME.

% \begin{figure}
%     \centering
%     \includegraphics[width=\linewidth]{images/Figure4.pdf}
%     \caption{\textbf{$\text{FP}{\rightarrow}\text{LIME}$.} If we make it so that FP distinguishes between forward and backward perturbations and scales attributions according to their distance from the input point, we can get results similar or even identical to LIME.}
%     \label{fig:fig4}
% \end{figure}
\begin{figure}
    \centering
    \includegraphics[width=\linewidth]{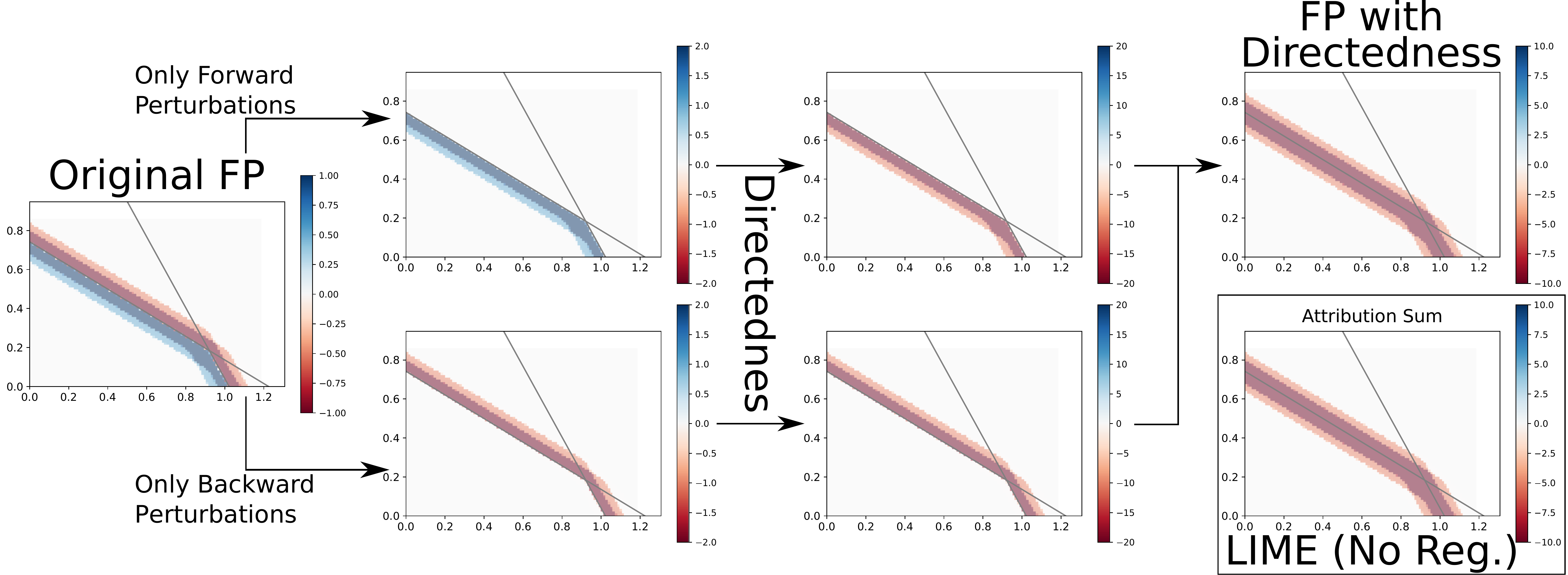}
    \caption{\textbf{$\text{FP}{\rightarrow}\text{LIME}$.} If we make it so that FP distinguishes between forward and backward perturbations and scales attributions according to their distance from the input point, we can get results similar or even identical to LIME.}
    \label{fig:fig4}
\end{figure}

An example of the FP relation to LIME can be seen in Figure~\ref{fig:fig4}. This is a simplified example with only four perturbations, each of which only changes a single feature by $0.1$. For this special case, we even reach identical attributions and while this is not generally true, the difference of LIME and FP after this transformation is also barely visible in most other cases. Also note that we used Linear Regression instead of Ridge Regression to easier show the equality after the transformation. 
% With Ridge Regression, the colormap stays the same but the attribution values are all closer to 0.

\section{Conclusions and Future Work}
We investigated the question of whether, and if so how, common attribution methods from XAI literature could be applied to settings beyond neural networks. Specifically, we looked at linear programs where we introduce various, sensible neural-encodings for representing the original LP.
% We show that attribution methods give meaningful results when applied on such NNs.
We show LIME and Saliency have very similar results if small perturbations are used for LIME.
By introducing the property of Directedness, we also found out how a perturbation-based Feature Attribution approach can be transformed to behave very similarly to LIME and, hence (for small perturbations), also to Saliency.
We believe that the discriminative properties (Tab.~\ref{tab:properties}) can guide the development of transparent and understandable attribution methods, while also paving the road for more general application in machine learning.

For future work, we can consider explaining special types of LPs as those used for quantifying uncertainty such as in MAP inference \citep{weiss2012map}. The application to mixed-integer LPs or integer LPs such as Shortest Path or Linear Assignment might prove valuable. Furthermore, investigating the scaling behavior of explanations generated for LPs or the cognitive aspects of whether and how LP attributions could be more ``human understandable'' seems critical.

\bibliographystyle{plainnat}
\subsubsection*{Acknowledgments}
This work was supported by the ICT-48 Network of AI Research Excellence Center ``TAILOR'' (EU Horizon 2020, GA No 952215), the Nexplore Collaboration Lab ``AI in Construction'' (AICO) and by the Federal Ministry of Education and Research (BMBF; project ``PlexPlain'', FKZ 01IS19081). It benefited from the Hessian research priority programme LOEWE within the project WhiteBox and the HMWK cluster project ``The Third Wave of AI'' (3AI).
\bibliography{neurips2022}

\begin{thebibliography}{22}
\providecommand{\natexlab}[1]{#1}
\providecommand{\url}[1]{\texttt{#1}}
\expandafter\ifx\csname urlstyle\endcsname\relax
  \providecommand{\doi}[1]{doi: #1}\else
  \providecommand{\doi}{doi: \begingroup \urlstyle{rm}\Url}\fi

\bibitem[Ancona et~al.(2019)Ancona, Ceolini, {\"O}ztireli, and
  Gross]{ancona2019gradient}
Marco Ancona, Enea Ceolini, Cengiz {\"O}ztireli, and Markus Gross.
\newblock Gradient-based attribution methods.
\newblock In \emph{Explainable AI: interpreting, explaining and visualizing
  deep learning}, pages 169--191. Springer, 2019.

\bibitem[Bazaraa et~al.(2008)Bazaraa, Jarvis, and Sherali]{bazaraa2008linear}
Mokhtar~S Bazaraa, John~J Jarvis, and Hanif~D Sherali.
\newblock \emph{Linear programming and network flows}.
\newblock John Wiley \& Sons, 2008.

\bibitem[Breiman(2001)]{breiman2001random}
Leo Breiman.
\newblock Random forests.
\newblock \emph{Machine learning}, pages 5--32, 2001.

\bibitem[Chinneck(2007)]{chinneck2007feasibility}
John~W Chinneck.
\newblock \emph{Feasibility and Infeasibility in Optimization:: Algorithms and
  Computational Methods}, volume 118.
\newblock Springer Science \& Business Media, 2007.

\bibitem[Ferber et~al.(2020)Ferber, Wilder, Dilkina, and
  Tambe]{ferber2020mipaal}
Aaron Ferber, Bryan Wilder, Bistra Dilkina, and Milind Tambe.
\newblock Mipaal: Mixed integer program as a layer.
\newblock In \emph{Proceedings of the AAAI Conference on Artificial
  Intelligence}, volume~34, pages 1504--1511, 2020.

\bibitem[Gunning and Aha(2019)]{gunning2019darpa}
David Gunning and David Aha.
\newblock Darpa’s explainable artificial intelligence (xai) program.
\newblock \emph{AI magazine}, pages 44--58, 2019.

\bibitem[Hesse et~al.(2021)Hesse, Schaub-Meyer, and Roth]{hesse2021fast}
Robin Hesse, Simone Schaub-Meyer, and Stefan Roth.
\newblock Fast axiomatic attribution for neural networks.
\newblock \emph{Advances in Neural Information Processing Systems}, 34, 2021.

\bibitem[Hoffman et~al.(1953)Hoffman, Mannos, Sokolowsky, and
  Wiegmann]{hoffman1953computational}
A~Hoffman, Murray Mannos, Daniel Sokolowsky, and N~Wiegmann.
\newblock Computational experience in solving linear programs.
\newblock \emph{Journal of the Society for Industrial and Applied Mathematics},
  pages 17--33, 1953.

\bibitem[Kokhlikyan et~al.(2020)Kokhlikyan, Miglani, Martin, Wang, Alsallakh,
  Reynolds, Melnikov, Kliushkina, Araya, Yan, and
  Reblitz-Richardson]{kokhlikyan2020captum}
Narine Kokhlikyan, Vivek Miglani, Miguel Martin, Edward Wang, Bilal Alsallakh,
  Jonathan Reynolds, Alexander Melnikov, Natalia Kliushkina, Carlos Araya, Siqi
  Yan, and Orion Reblitz-Richardson.
\newblock Captum: A unified and generic model interpretability library for
  pytorch, 2020.

\bibitem[Krizhevsky et~al.(2012)Krizhevsky, Sutskever, and
  Hinton]{krizhevsky2012imagenet}
Alex Krizhevsky, Ilya Sutskever, and Geoffrey~E Hinton.
\newblock Imagenet classification with deep convolutional neural networks.
\newblock \emph{NeurIPS}, 2012.

\bibitem[Mnih et~al.(2013)Mnih, Kavukcuoglu, Silver, Graves, Antonoglou,
  Wierstra, and Riedmiller]{mnih2013playing}
Volodymyr Mnih, Koray Kavukcuoglu, David Silver, Alex Graves, Ioannis
  Antonoglou, Daan Wierstra, and Martin Riedmiller.
\newblock Playing atari with deep reinforcement learning.
\newblock \emph{arXiv preprint arXiv:1312.5602}, 2013.

\bibitem[Paulus et~al.(2021)Paulus, Rol{\'\i}nek, Musil, Amos, and
  Martius]{paulus2021comboptnet}
Anselm Paulus, Michal Rol{\'\i}nek, V{\'\i}t Musil, Brandon Amos, and Georg
  Martius.
\newblock Comboptnet: Fit the right np-hard problem by learning integer
  programming constraints.
\newblock In \emph{International Conference on Machine Learning}, pages
  8443--8453. PMLR, 2021.

\bibitem[Ribeiro et~al.(2016)Ribeiro, Singh, and Guestrin]{ribeiro2016should}
Marco~Tulio Ribeiro, Sameer Singh, and Carlos Guestrin.
\newblock " why should i trust you?" explaining the predictions of any
  classifier.
\newblock In \emph{Proceedings of the 22nd ACM SIGKDD international conference
  on knowledge discovery and data mining}, pages 1135--1144, 2016.

\bibitem[Schaber et~al.(2012)Schaber, Steinke, and
  Hamacher]{schaber2012transmission}
Katrin Schaber, Florian Steinke, and Thomas Hamacher.
\newblock Transmission grid extensions for the integration of variable
  renewable energies in europe: Who benefits where?
\newblock \emph{Energy Policy}, 43:\penalty0 123--135, 2012.

\bibitem[Selvaraju et~al.(2017)Selvaraju, Cogswell, Das, Vedantam, Parikh, and
  Batra]{selvaraju2017grad}
Ramprasaath~R Selvaraju, Michael Cogswell, Abhishek Das, Ramakrishna Vedantam,
  Devi Parikh, and Dhruv Batra.
\newblock Grad-cam: Visual explanations from deep networks via gradient-based
  localization.
\newblock In \emph{Proceedings of the IEEE international conference on computer
  vision}, pages 618--626, 2017.

\bibitem[Simonyan et~al.(2014)Simonyan, Vedaldi, and
  Zisserman]{simonyan2014deep}
Karen Simonyan, Andrea Vedaldi, and Andrew Zisserman.
\newblock Deep inside convolutional networks: Visualising image classification
  models and saliency maps.
\newblock In \emph{In Workshop at International Conference on Learning
  Representations}. Citeseer, 2014.

\bibitem[Stammer et~al.(2021)Stammer, Schramowski, and
  Kersting]{stammer2021right}
Wolfgang Stammer, Patrick Schramowski, and Kristian Kersting.
\newblock Right for the right concept: Revising neuro-symbolic concepts by
  interacting with their explanations.
\newblock In \emph{Proceedings of the IEEE/CVF Conference on Computer Vision
  and Pattern Recognition}, pages 3619--3629, 2021.

\bibitem[Sundararajan et~al.(2017)Sundararajan, Taly, and
  Yan]{sundararajan2017axiomatic}
Mukund Sundararajan, Ankur Taly, and Qiqi Yan.
\newblock Axiomatic attribution for deep networks.
\newblock In \emph{International Conference on Machine Learning}, pages
  3319--3328. PMLR, 2017.

\bibitem[Teso and Kersting(2019)]{teso2019explanatory}
Stefano Teso and Kristian Kersting.
\newblock Explanatory interactive machine learning.
\newblock In \emph{Proceedings of the 2019 AAAI/ACM Conference on AI, Ethics,
  and Society}, pages 239--245, 2019.

\bibitem[Vaswani et~al.(2017)Vaswani, Shazeer, Parmar, Uszkoreit, Jones, Gomez,
  Kaiser, and Polosukhin]{vaswani2017attention}
Ashish Vaswani, Noam Shazeer, Niki Parmar, Jakob Uszkoreit, Llion Jones,
  Aidan~N. Gomez, Lukasz Kaiser, and Illia Polosukhin.
\newblock Attention is all you need.
\newblock In \emph{NeurIPS}, 2017.

\bibitem[Weiss et~al.(2007)Weiss, Yanover, and Meltzer]{weiss2012map}
Yair Weiss, Chen Yanover, and Talya Meltzer.
\newblock Map estimation, linear programming and belief propagation with convex
  free energies.
\newblock \emph{UAI}, 2007.

\bibitem[Zeiler and Fergus(2014)]{zeiler2014visualizing}
Matthew~D Zeiler and Rob Fergus.
\newblock Visualizing and understanding convolutional networks.
\newblock In \emph{European conference on computer vision}, pages 818--833,
  2014.

\end{thebibliography}

\clearpage
\appendix

\section{Appendix for ``Attributions Beyond Neural Networks: The Linear Program Case''}
This appendix contains an example of single feature attributions for the experiment shown in Figure~\ref{fig:results} and some technical details.

\subsection{Single Feature Attributions}
\label{app:feature_attr}

\begin{figure*}[ht]
    \centering
    \includegraphics[width=\linewidth,origin=c]{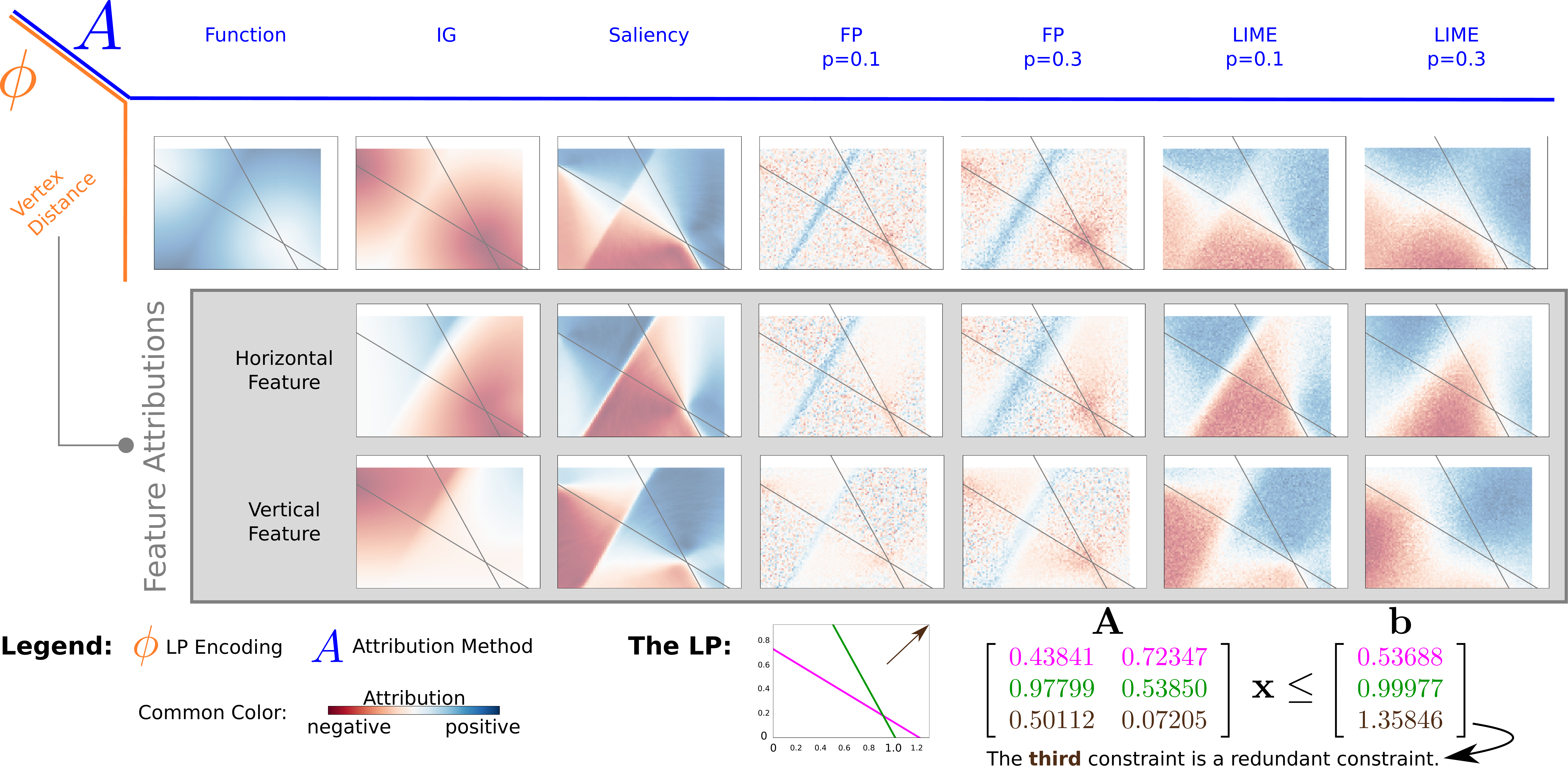}
    % \vspace{.25cm}
    \caption{\textbf{Single feature attributions for the Vertex Distance encoding.}
    The data is from the same experiment as in Figure~\ref{fig:results}.
    The LP has two features, one horizontal and one vertical, where the grey lines indicate the constraints for that LP. The line on the top shows the summed up attribution of all features for the Vertex Distance encoding. Each column shows a different attribution method (or configuration). For FP and LIME, $\text{p}$ indicates the maximum possible perturbation in any direction. The (rounded) numbers for the constraints of that LP are shown on the bottom right. The part enclosed in the grey box titled ``Feature Attributions'' shows attribution for both single features on the Vertex Distance encoding.
    %The constraint of the third row in $\mathbf{A}$ is a redundant constraint, it does not restrict the set of feasible instances.
    (Best viewed in color.)
    }
    \label{fig:appendix}
\end{figure*}

In Figure~\ref{fig:appendix}, feature attributions for both features on the example of the Vertex Distance encoding are shown.
These plots have been created using the same experiment as in Figure~\ref{fig:results}.
Generally, it can be seen how the methods also differ in their single feature attributions and that feature attributions can vary significantly depending on the respective feature.
The goal of this section is to inspect the results shown in Figure~\ref{fig:appendix} to go into detail about the behavior of the attribution methods used in this paper and to explain the differences between the attributions for the two features here.\footnote{Readers which are very familiar with these methods might find large parts of the following explanations obvious, for less versed readers, these explanations hopefully can help to make the results more understandable.}

\textbf{Integrated Gradients.}
Attribution for IG is obtained by calculating gradients on the path from a baseline to the input vector.
As mentioned in the paper, this experiment uses the origin as its baseline: $(0, 0)$.
To illustrate this calculation, consider points which only differ from the baseline on the vertical feature (the far left of the shown plots).
Here, the attribution for the horizontal feature remains at $0$, as there is no change of this feature compared to the baseline, so it can not be responsible for any change in the output.
Moving up towards the top left vertex however, the vertical feature gets an increasingly larger negative attribution.
Since the attributions of IG sum up to the difference in output between baseline and input instance (Completeness), this difference has to be reflected by the feature attributions.
In this encoding, the function values encode the distance to the nearest vertex.
The encoding at the baseline (origin) has some positive value and the lowest values (the smallest possible distance, i.e. $0$) can be found on any of the vertices.
Therefore, the attribution for the vertical feature on the left vertex (where the horizontal feature remains unchanged compared to the baseline) must equal the negative difference between baseline output and $0$, since otherwise the Completeness property would be violated.
Another interesting region in Figure~\ref{fig:appendix} can be found on the top, roughly in the middle of our plot.
Here is the same kind of negative attribution for the vertical feature but also some positive attribution for the horizontal feature.
%Similar to the previous explanation, the negative attribution is there because having a larger vertical feature value gets the point closer to the vertex (where the output, i.e. the distance to a vertex is smaller), but now, in that area with positive horizontal attribution, the output would be smaller if the point was not so far on the right.
While the increased vertical feature decreases the distance to that vertex, the increased horizontal feature increases the distance which results in positive attribution for that feature.
In other words, the horizontal feature being larger than its baseline value ($0$) has a positive (the distance increases) impact on the output.
This same reasoning also applies to other regions but there, the attribution can be influenced by other vertices, resulting in different attributions. For example, being on the bottom right indicates a small distance to the nearest vertex there (one of the two vertices on the bottom right), which is why for such instances positive horizontal feature values can have negative attributions.

\textbf{Saliency.}
Since Saliency  uses the local gradient for its attributions, understanding gradients is mostly sufficient for understanding this attribution method. 
To put it simply, on a specific point, if a feature has a positive impact on the output, then the gradient is positive, and if a feature has a negative impact on the output, then the gradient is negative.
This impact and therewith the gradient can also be $0$. 
Let us consider the Saliency feature attributions in Figure~\ref{fig:appendix}.
In accordance with Directedness, if increasing a feature would get the point closer to a vertex (decrease the distance), it gets negative attribution.
If increasing a feature would get the point further away from a vertex (increase the distance), it gets positive attribution.\footnote{Keep in mind that, like with gradients, such statements of ``increasing'' and ``decreasing'' should not be interpreted with some specific amount of change but rather infinitesimal changes of the input.}
The white areas in this figure might benefit from some further explaining.
First of all, on and around the vertices, there is (close to) $0$ attribution.
The vertices are points where in theory the gradient should be undefined (because of the absolute value function used in the encoding).
However, the NN approximates the true underlying function in a continuous way, leading to a continuously changing gradient and an attribution of $0$ on the vertices which rather quickly reaches the ``normal'' gradient when moving away from those vertices in either direction.
The white line starting at the bottom (somewhat left) and ending on the top (somewhat right) also has many points with (close to) $0$ attribution because in this region, the gradient switches signs.
For the single feature attributions, there are also some white lines moving away from the vertices: to the top/bottom for the horizontal feature and to the left/right for the vertical feature.
Those are areas where only that respective feature changes its gradient from one sign to another.
For example, if a point is below the nearest vertex, then its vertical feature attribution is negative first and once the point is above the closest vertex, the attribution becomes positive.
At some point in between, when the point is on the same height as the vertex, its feature attribution is 0.
This is true both if these points are exactly under/above the vertex or slightly on one side.
Combining this behavior for points on the left and right of the vertices results in a horizontal white line indicating no attribution next to the vertex.
The explanation for the horizontal feature can be done the same way.
Such white lines resulting from this type of behavior are not necessarily visible in the plot for the attribution sum, since here the other features can have non-zero attributions.
However, the attribution sum can also contain areas with no attribution even though there are single feature attributions because in such situations, these get averaged out to $0$ overall (this can also be seen in Figure~\ref{fig:appendix}).

\textbf{Feature Permutation.}
The attributions of FP need to be interpreted differently as they are unlike the other attribution methods in Figure~\ref{fig:appendix}.
For FP, the output of the input point is compared to the outputs of neighboring points (perturbations).
If, on average, their output is larger, then the attribution of the input instance is negative which describes the behavior seen around the vertices.
If the neighboring points are, on average, smaller, then the attribution is positive.
For the Vertex Distance encoding, this blue (positive) attribution can be observed on the line which has an equal distance to both nearest vertices, as here perturbing any feature in any direction creates a point closer to a vertex.
Note, that FP always considers only one feature changed at a time since perturbations are only created by perturbing a single feature.
The single feature attributions also indicate how important the features are compared to each other.
For example, on this aforementioned blue line, the attribution is larger for the horizontal feature.
This makes sense, as moving to the left or right has a higher impact than moving up or down by the same distance.\footnote{Because the line is more vertical than horizontal, changing the horizontal feature can get a point closer to a vertex. In other words, since the vertical distance between the two relevant vertices is smaller than the horizontal distance, changing the horizontal distance here has a higher impact on the output.}
% Comparing both features, the horizontal features seems to have a stronger impact overall.
% This makes sense, as the vertical distance between the two vertices which determines the majority of the encoding here\footnote{The vertex on the far bottom (on the right) is less important than the other two vertices because, compared to the vertex on top of it, there are less points which are closest to it. The vertex on the left also has many points to which it is the closest vertex.} is a bit smaller than the horizontal distance, therefore any perturbation of the same scale tends to have a larger impact on the horizontal feature. % TODO explain better?
The noise visible in many regions is a result of the randomness in the FP perturbations.
For example, in many areas the score increases roughly the same in one direction as it decreases in the other direction, therefore the FP attribution should be around $0$.
But if due to randomness, perturbations are created more strongly in one direction, the average output is now predominantly influenced by that direction, resulting in an average positive or negative change and an attribution not close to $0$.
This can happen in either direction, which in Figure~\ref{fig:appendix} results in noisy areas with many red and blue dots in an otherwise rather white region.
Interestingly enough, the strength of this noise for both features here indicates in which areas which feature has a higher impact (this even compares to the respective Saliency feature attributions).
The larger the perturbations (p in Figure~\ref{fig:appendix}), the broader the area considered for the attributions, resulting in less accurate (local) but increasingly robust attributions that represent more general changes.

\textbf{LIME.}
The similarity between LIME and Saliency is not only present for the attribution sum but also for the single feature attributions.
Instead of using the local gradient, LIME creates a small model for the input point based on perturbed instances around it.
To summarize the resulting attributions briefly:
feature attribution for LIME is positive if larger (smaller) instances have larger (smaller) outputs, negative if larger (smaller) instances have smaller (larger) scores and zero if either their outputs are smaller in one but equally larger in the other direction or if instances around the input have the same output as the input itself.
Note, that, as with Saliency, the direction of the change around the input instance matters (Directedness).
Also keep in mind that, unlike FP, LIME uses perturbations for multiple features at a time.
For larger perturbations, it can be seen that the attribution patterns become more blurred and local details are disappearing.
In some situations, this could possibly be an advantage and protect against noise, making the results more robust.
There can be some slightly differing attributions of similar points due to the randomness in the perturbations.

\subsection{Technical Details}
Our code is available at \url{https://cutt.ly/gHE0Nkx}.
All experiments were run on a desktop machine with an Intel(R) Core(TM) i5-7500 CPU processor, 16GB RAM, and a NVIDIA GeForce GTX 1070 graphics card.

\end{document}